\pdfoutput=1

\documentclass[11pt]{article}

\usepackage[final]{coling}

\usepackage{times}
\usepackage{latexsym}

\usepackage[T1]{fontenc}

\usepackage[utf8]{inputenc}

\usepackage{microtype}

\usepackage{inconsolata}

\usepackage{graphicx}

\usepackage{rotating}

\usepackage{listings}

\usepackage{float}

\title{ESNLIR: A Spanish Multi-Genre Dataset with Causal Relationships}

\author{Johan Rodríguez \and Nicolás Perez \and Rubén Manrique \\
         jd.rodriguezp1234@uniandes.edu.co \\ n.perezt2@uniandes.edu.co \\ rf.manrique@uniandes.edu.co}

\begin{document}
\maketitle
\nolinenumbers
\begin{abstract}
Natural Language Inference (NLI), also known as Recognizing Textual Entailment (RTE), serves as a crucial area within the domain of Natural Language Processing (NLP). This area fundamentally empowers machines to discern semantic relationships between assorted sections of text. Even though considerable work has been executed for the English language, it has been observed that efforts for the \textbf{S}panish language are relatively sparse. Keeping this in view, this paper focuses on generating a multi-genre Spanish dataset for \textbf{NLI}, \textbf{ESNLIR}, particularly accounting for causal \textbf{R}elationships. A preliminary baseline has been conceptualized and subjected to an evaluation, leveraging models drawn from the BERT family. The findings signify that the enrichment of genres essentially contributes to the enrichment of the model's capability to generalize.

The code, notebooks and whole datasets for this experiments is available at: \url{https://zenodo.org/records/15002575}. If you are interested only in the dataset you can find it here: \url{https://zenodo.org/records/15002371}. 
\end{abstract}

\section{Introduction}

In the field of Natural Language Processing, several applications including Information Retrieval (IR), Question Answering (QA), and Information Extraction (IE) necessitate machine comprehension of semantic meaning in text corpora \cite{daganpPascal}. However, understanding isolated sentences is insufficient in these contexts. Being able to grasp the relationships between sentences is equally imperative. Consequently, Natural Language Inference (NLI) or Recognizing Textual Entailment (RTE) has been developed to ascertain whether one sentence, the hypothesis, can be inferred from a different sentence, the premise \cite{daganpPascal}. Nonetheless, a binary label proved inadequate for encompassing all possible semantic relationships between sentences, such as contradiction \cite{rte4}. Hence, a three-way labeling scheme was devised, which comprises of \textit{Entailment}, \textit{Neutral} and \textit{Contradiction} classes.

This labeling technique has been employed in the development of various contemporary benchmarks such as RTE4 \cite{rte4}, SICK \cite{SICK}, which are small datasets with less than 10,000 examples,  SNLI \cite{SNLI}, the first large scale dataset with more than 500,000 examples, MNLI \cite{MNLI}, the multi-genre version of SNLI, and XNLI \cite{XNLI}, a multilingual subset of XNLI. While commendable classification performance has been achieved on these benchmarks, the sciNLI \cite{SCINLI} benchmark underlines the need for a novel label, \textit{Reasoning}, to encapsulate causal relationships in genres such as scientific text, where ideas often manifest as cause-effect sequences. The same benchmark introduces a method for automated premise-hypothesis pair extraction, diverging from the standard practice of requiring human annotation for sentence generation, thus facilitating the extraction of large volumes of examples without human supervision. This method has also been successfully employed in the msciNLI benchmark \cite{MSCINLI} for multi-genre configurations. In particular, these works \cite{SCINLI, MSCINLI} state that linking phrases between sentences are a consequence of their type of semantic relationship, and that some linking phrases such as "thus" and "therefore" show cause-effect relationships between sentences that most previous works did not take into account.

However, limited research has been executed on NLI in Spanish, with the exception of the Spanish corpus included in XNLI. The first benchmark in this area was created by generating sentence templates from a Spanish Question Answering corpus \cite{SPARTE} but is relatively small having 2,962 examples and adopts a binary label for \textit{Entailment}. Despite being subjected to traditional models such as Support Vector Machines, the SPARTE benchmark achieved an accuracy of 60.83 \cite{Castillo2010}.  Another benchmark, INFERES \cite{INFERES}, produced with human annotation and exhibiting a three-way labeling technique, scans across six domains extracted from Wikipedia. It is rather more extensive with 8,055 examples and includes a \textit{Contradiction} category. Other studies \cite{transwordses} have indicated that viable results can be attained in medical domain text by employing linking phrases to extract sentence pairs and ensuing classification of sentence relationships. Additionally, certain model-based approaches like SILT \cite{SILT} and Facter-Check \cite{facter-check} have demonstrated that BERT-based models can achieve good classification performance. However, these models may unintentionally learn heuristics from annotation artifacts \cite{rightforwrong} and may even predict labels by merely examining the hypothesis \cite{hippopred}.

It is apparent from the existing Spanish NLI research that there is significant room for improvement. One avenue could be the creation of a larger benchmark to facilitate the training of expansive deep learning architectures. Ideally, such a benchmark would span multiple genres, thereby encompassing various writing styles and multitudinous topics to enhance model generalization, akin to MNLI and MSciNLI. Further improvements could be made by including a label to account for causal relationships, as exhibited in sciNLI and MSciNLI. Not least, model evaluations should seek to measure vulnerability to learning artificial patterns from the annotation artifacts in the benchmark in addition to conventional classification metrics \cite{hippopred} \cite{naik-etal-2018-stress}. In light of these considerations, we propose a novel Spanish NLI multi-genre dataset, \textbf{ESNLIR}, Reasoning-Spanish-Natural-Language-Inference dataset, the extraction methodology for which is based on sciNLI. We apply BERT-based models to corpora of multiple genres to establish a baseline of not only classification metrics but also a method for detecting potential annotation and model artifacts.

With this in mind, this paper first defines the dataset extraction, followed by the training and evaluation results for the baseline and BERT-based models, including insights into the model annotation artefacts and stress tests.

\section{Dataset}

The data collection process for ESNLIR is akin to that outlined in sciNLI, focusing on searching for linking phrases within extensive corpora. This approach allows for the automatic extraction of premise-hypothesis pairs, accompanied by a brief label validation procedure to ensure a basic standard of data quality. In this section, we provide a detailed description of this process, covering the sources of corpora, genres, and cardinality.

\subsection{Multi-Genre Corpora}
Following the hypothesis from the MNLI dataset, which suggests that incorporating multiple genres enhances models' generalization capabilities, we selected 34 Spanish corpora to represent eight distinct writing genres:
\begin{itemize}
    \item articles: Web articles and encyclopedia articles.
    \item books: Literature in general, consisting of open access or public domain books.
    \item comments: Web comments on social networks, accounting for non-formal texts.
    \item legal: Legal documents from Colombia.
    \item clinical: Clinical cases extracted from open source datasets.
    \item news: News articles extracted from multiple sources.
    \item talks: Speeches extracted from TED, which extend the non-formal texts group.
    \item theses: Theses extracted from the website of the Universidad de Los Andes, Colombia, with access for academic purposes.
\end{itemize}
The domain and genre of each selected corpora can be found in Table \ref{tab:split_count}.

\subsection{Premise-hypothesis extraction}
For each corpus, an identical process is used to extract premise-hypothesis pairs. Using a methodology similar to sciNLI, certain linking phrases are identified as indicators of various semantic relationships between sentences, as demonstrated in Table \ref{tab:classes}. Based on these criteria, for each document in the corpus, sentences starting with any of the linking phrases listed in Table \ref{tab:classes} are identified and used as hypotheses in sentence pairs. To construct these sentence pairs, the sentences that end just before the linking phrases and are located in the same paragraph are taken as premises. Ultimately, the label for each premise-hypothesis pair is determined based on the label allocated to the linking phrase in Table \ref{tab:classes}. In executing this process, both premises and hypotheses for examples from the \textbf{contrasting}, \textbf{entailment}, and \textbf{reasoning} categories are extracted from the same paragraph. Conversely, neutral examples are created by pairing sentences that do not fall into the aforementioned categories and originate from different paragraphs. Specifically, there are multiple methods to create \textbf{neutral} examples:
\begin{itemize}
    \item Both random: Pairing of sentences not previously used as premises or hypotheses in the other classes.
    \item First random: The first sentence is a previously used premise and the hypothesis is a random, previously unused sentence.
    \item Second random: The first sentence is a previously unused random sentence and the second is a previously used hypothesis.
    \item Both entailed: Both premise and hypothesis have been used previously in other pairs, but belong to different paragraphs in the same document.
\end{itemize}

Since the neutral class has a completely different extraction method, which mainly differs in taking text fragments from different paragraphs and not using linking phrases, it is expected that the classifiers will have some bias towards this class, causing it to have better classification metrics than the others.

To illustrate the process, in the following paragraph taken from Wikipedia, \textbf{Sin embargo} is a linking phrase belonging to the \textbf{contrasting} class:

\textbf{Text in Spanish}: 

\textit{"Formado en el diario El País de Madrid, fundó y dirigió el periódico Siglo 21 de Guadalajara desde su fundación en 1991 hasta el 30 de abril de 1997, día en el que renunció, para crear el diario Público. \textcolor{blue}{Un día antes de que Zepeda anunció el primer número de Público, hubo una operación que buscó cerrar a Siglo 21, mediante la salida masiva de sus empleados}. \textbf{Sin embargo}, \textcolor{red}{Siglo 21 sobrevivió a ese intento de desaparecerlo y, finalmente, Zepeda se vio obligado a vender 66.66 por ciento de Público a los propietarios de Grupo Multimedios, (que posteriormente renombraron a Público como Milenio Jalisco, denominación que mantiene hasta la fecha)}. En 1999 deja Público para asumir la subdirección de El Universal en la Ciudad de México, cargo que desempeñaría hasta 2001."}

\textbf{Text in English}
\textit{"Trained at El País newspaper in Madrid, he founded and directed the newspaper Siglo 21 in Guadalajara from its founding in 1991 until April 30, 1997, the day he resigned to create the newspaper Público. \textcolor{blue}{A day before Zepeda announced the first issue of Público, there was an operation that sought to close Siglo 21, through the massive departure of its employees}. \textbf{However}, \textcolor{red}{Siglo 21 survived that attempt to disappear it and, finally, Zepeda was forced to sell 66.66 percent of Público to the owners of Grupo Multimedios, (who subsequently renamed Público as Milenio Jalisco, a denomination it maintains to date)}. In 1999, he left Público to take over as assistant editor of El Universal in Mexico City, a position he held until 2001."}

In this context, the sentence highlighted in blue can be considered as the premise, and the sentence highlighted in red as the hypothesis. The premise discusses an operation aimed at closing the magazine Siglo 21, whereas the hypothesis \textbf{contrasts} this by stating that the magazine ultimately survived.

After extraction, sentences are refined by retaining only those that are syntactically complete. To achieve this, Part-of-Speech (POS) tagging is performed on each sentence using the Spacy es\_core\_news\_lg model. Sentences are preserved only if they contain both a subject (with tags such as "NOUN", "PRON", "PROPN") and a predicate (with tags such as "AUX", "VERB").

\subsection{Train-Val-Test split}
In order to facilitate the evaluation of the dataset, for each corpus all splits are extracted in a balanced way. In addition, to avoid sharing pairs between splits, a double stratification strategy is used to generate the splits:
\begin{enumerate}
    \item Separate articles into groups, one for each split, to avoid shared articles between splits. First a training and validation vs testing split is made, then a training  vs validation split completes the grouping.
    \item For each split find the class with the minimum number of examples and downsample the rest of the classes to have the same number of examples.
\end{enumerate}

As shown in table \ref{tab:split_count}, a maximum of 15000 examples have been selected for testing and validation splits, but the balance of classes is maintained for all splits. However, there is some imbalance in the genre representation, which is caused by the difference in the number of articles found in each of the corpora. Apart from that, some corpora were selected to be used in the test only, in order to test if the dataset offers some form of domain generalisation. In particular, the following corpora were selected for testing only for the following reasons:
\begin{itemize}
    \item esbooks\_\_eswikibooks and esbooks\_\_libreriaunal: It is expected that the semantic relations learnt in other book corpora, i.e. from traditional literature, can help to learn fenomena in web books and college books such as those found in wikibooks and libreria unal respectively.
    \item escomments\_\_reddit and escomments\_\_suicide: It is expected that relations learned from Twitter comments will help to learn behaviours in these corpora.
    \item eslegal\_\_entrevistas\_comision\_verdad: Since these interviews are informal, it is interesting to check whether the models are able to learn this type of relations from formal sources on the same topics.
    \item esmedical\_\_BVS and esmedical\_\_SPACC and esmedical\_\_TEI\_ES: As the training split contains information about medical theses, it is expected that the models can generalise to these corpora.
    \item esnews\_\_eswikinews: Similar to esbooks\_\_eswikibooks, it is expected that learning from traditional news sources can help to predict web news.
    \item estalks\_\_TED: It is expected that a combination of non-formal sources such as twitter and all the domains presented in the rest of the training corpora can help to predict relations in sentences belonging to talks.
\end{itemize}
Finally, for each split all the corpora examples are merged into a single split, resulting in a single dataset with 7'325.356 training examples, 127.404 validation examples, and 128.412 test examples.

\begin{sidewaystable*}[]
\centering
\resizebox{\columnwidth}{!}{
\begin{tabular}{lllllll}
\hline
\multicolumn{1}{|c|}{\textbf{dataset}}                             & \multicolumn{1}{c|}{\textbf{train|unique\_articles}} & \multicolumn{1}{c|}{\textbf{train|total\_examples}} & \multicolumn{1}{c|}{\textbf{val|unique\_articles}} & \multicolumn{1}{c|}{\textbf{val|total\_examples}} & \multicolumn{1}{c|}{\textbf{test|unique\_articles}} & \multicolumn{1}{c|}{\textbf{test|total\_examples}} \\ \hline
esarticles\_\_eswiki                                               & 75004                                                & 254756                                              & 1907                                               & 5976                                              & 2051                                                & 6568                                               \\
esbooks\_\_crawling                                                & 75122                                                & 2653328                                             & 209                                                & 6848                                              & 197                                                 & 7448                                               \\
esbooks\_\_elchico                                                 & 582                                                  & 9600                                                & 69                                                 & 696                                               & 91                                                  & 1828                                               \\
esbooks\_\_eltec                                                   & 37                                                   & 620                                                 & 10                                                 & 124                                               & 8                                                   & 100                                                \\
esbooks\_\_eswikibooks                                             & 0                                                    & 0                                                   & 0                                                  & 0                                                 & 135                                                 & 672                                                \\
esbooks\_\_gutenberg                                               & 455                                                  & 7404                                                & 44                                                 & 776                                               & 55                                                  & 904                                                \\
esbooks\_\_libreriaunal                                            & 0                                                    & 0                                                   & 0                                                  & 0                                                 & 402                                                 & 6588                                               \\
escomments\_\_reddit                                               & 0                                                    & 0                                                   & 0                                                  & 0                                                 & 2009                                                & 2860                                               \\
escomments\_\_suicide                                              & 0                                                    & 0                                                   & 0                                                  & 0                                                 & 25                                                  & 36                                                 \\
escomments\_\_tweets                                               & 52573                                                & 66708                                               & 6463                                               & 8176                                              & 6504                                                & 8244                                               \\
eslegal\_\_entrevistas\_comision\_verdad                           & 0                                                    & 0                                                   & 0                                                  & 0                                                 & 8                                                   & 68                                                 \\
eslegal\_\_informes\_analisis\_comision\_verdad                    & 3                                                    & 52                                                  & 1                                                  & 4                                                 & 1                                                   & 4                                                  \\
eslegal\_\_raw\_sentencias\_corte\_colombia                        & 18020                                                & 249424                                              & 385                                                & 7020                                              & 407                                                 & 5556                                               \\
esmedical\_\_BVS                                                   & 0                                                    & 0                                                   & 0                                                  & 0                                                 & 12                                                  & 16                                                 \\
esmedical\_\_TEI\_ES                                               & 0                                                    & 0                                                   & 0                                                  & 0                                                 & 89                                                  & 180                                                \\
esnews\_\_colombian\_news                                          & 8043                                                 & 18272                                               & 1134                                               & 2268                                              & 1140                                                & 2264                                               \\
esnews\_\_eswikinews                                               & 0                                                    & 0                                                   & 0                                                  & 0                                                 & 7                                                   & 12                                                 \\
esnews\_\_spanish\_pd\_news                                        & 207471                                               & 858268                                              & 1971                                               & 7460                                              & 1991                                                & 7972                                               \\
estalks\_\_TED                                                     & 0                                                    & 0                                                   & 0                                                  & 0                                                 & 156                                                 & 548                                                \\
estheses\_\_Centro Interdisciplinario de Estudios sobre Desarrollo & 320                                                  & 5460                                                & 26                                                 & 380                                               & 26                                                  & 520                                                \\
estheses\_\_Escuela de Gobierno Alberto Lleras Camargo             & 232                                                  & 3308                                                & 21                                                 & 276                                               & 22                                                  & 440                                                \\
estheses\_\_Facultad de Administración                             & 622                                                  & 11664                                               & 56                                                 & 944                                               & 53                                                  & 696                                                \\
estheses\_\_Facultad de Arquitectura y Diseño                      & 1204                                                 & 12700                                               & 139                                                & 1444                                              & 131                                                 & 1444                                               \\
estheses\_\_Facultad de Arte y Humanidades                         & 623                                                  & 11120                                               & 78                                                 & 1200                                              & 63                                                  & 912                                                \\
estheses\_\_Facultad de Ciencias                                   & 985                                                  & 16036                                               & 123                                                & 1960                                              & 105                                                 & 2224                                               \\
estheses\_\_Facultad de Ciencias Sociales                          & 1676                                                 & 50872                                               & 188                                                & 5544                                              & 207                                                 & 6452                                               \\
estheses\_\_Facultad de Derecho                                    & 1600                                                 & 32412                                               & 199                                                & 3908                                              & 181                                                 & 3968                                               \\
estheses\_\_Facultad de Economía                                   & 1201                                                 & 17828                                               & 149                                                & 2112                                              & 150                                                 & 1980                                               \\
estheses\_\_Facultad de Educación                                  & 671                                                  & 16264                                               & 73                                                 & 2152                                              & 75                                                  & 1944                                               \\
estheses\_\_Facultad de Ingeniería                                 & 5660                                                 & 83284                                               & 503                                                & 7236                                              & 493                                                 & 6904                                               \\
estheses\_\_Facultad de Medicina                                   & 19                                                   & 132                                                 & 4                                                  & 16                                                & 4                                                   & 28                                                 \\
estheses\_\_Unknown                                                & 507                                                  & 8456                                                & 65                                                 & 880                                               & 66                                                  & 836                                                \\
                                                                   &                                                      &                                                     &                                                    &                                                   &                                                     &                                                    \\
                                                                   &                                                      &                                                     &                                                    &                                                   &                                                     &                                                    \\
                                                                   &                                                      &                                                     &                                                    &                                                   &                                                     &                                                   
\end{tabular}
}
\caption{Corpora split example count. Some of the corpora is out-of-sample, meaning that it does not have examples in training and validation splits}
\label{tab:split_count}
\end{sidewaystable*}

\begin{table*}[]                                     
\centering
\small
\begin{tabular}{p{0.2\linewidth} | p{0.4\linewidth} | p{0.4\linewidth}}
\hline
Class       & Description & Linking phrases
\\
\hline

contrasting & The hypothesis contradicts or mentions a comparison, criticism, juxtaposition, or a limitation of something said in the premise. & "sin embargo," "no obstante," "por otra parte,"  "por otro lado," "en cambio," "por el contrario," "al contrario," "en contraste,"                                                                                                                                                                                                                                                                                                                                                                                                                                                                                                                                                                                                                                                                                                                                                                                                                                               \\
\hline
entailment  & The hypothesis generalizes, specifies or has an equivalent meaning with the premise.                                             & "en concreto," "concretamente,"  "especificamente," "precisamente," "en particular," "particularmente," "en especial," "es decir," "en otras palabras," "dicho de otra manera," "dicho de otro modo," "en otros terminos," "de hecho," "esto es," "o sea," "mejor dicho," "sobre todo," "justamente,"   "en resumidas cuentas," "en resumen," "en breve," "por ejemplo," "en sintesis," "en efecto," "en pocas palabras," "en una palabra," "recapitulando," "brevemente,"  "recogiendo lo mas importante," "como se ha dicho," "para ilustrar,"
\\
\hline
neutral     & Premise and hypothesis are semantically independent of each other.                                                               &                                                                                                                                                                                                                                                                                                                                                                                                                                                                                                                                                                                                                                                                                                                                                                                                                                                                                                                                                                                                                                                                                                                         \\
\hline
reasoning   & The premise presents the reason, cause, or condition for the result or conclusion made in the hypothesis.                        & "por lo tanto," "por tanto," "en consecuencia," "por consiguiente," "por ende," "por esa razon," "por eso," "de ahi que," "como resultado," "como consecuencia,"                                                                                                                                                                                                                    \\                                                     \hline                                                                                                                                                                                                                                                                                                                                                                                                                                     
\end{tabular}
\caption{Classes and linking phrases used to group different relation types between premise and hypothesis pairs.}
\label{tab:classes} 
\end{table*}

\section{Experimental setup}
This section describes the experimental setup used to train and evaluate the performance of several models, and the dataset in general.

\subsection{Models}
The following models have been selected to provide a performance baseline for further research:
\begin{itemize}
    \item XGBoost \cite{xgboost}: This is an ensemble of three models used as a baseline for lexical representations, in fact a Bag Of Words representation, to determine whether lexical representations are sufficient to classify the examples. For this model, 2000 estimators were used.
    \item BERTIN \cite{BERTIN}: A BERT-based architecture that removes the next sentence prediction task. Specifically, it uses the Spanish pre-trained model bertin-project/bertin-roberta-base-spanish from huggingfaces.
    \item XLMRoBERTa \cite{xlmroberta}: A multilingual version of RoBERTa to test if our dataset performs well on multilingual configurations. This pre-trained model is also extracted from huggingfaces, with the path FacebookAI/xlm-roberta-base.
\end{itemize}

\subsection{Metrics}
As the test dataset is balanced, the accuracy and f1\_score macro are used to evaluate the performance of the models.

\subsection{Training and Fine-Tuning}
The XGBoost model was trained from scratch using a Bag Of Words representation, while the BERT-based models were fine-tuned from the pre-trained models up to a maximum of 6 epochs, with early stopping via the f1\_score metric, a batch size of 64 and a learning rate of 2e-5.

\subsection{Dataset artifact detection}
To detect dataset artifacts the strategy applied in \cite{hippopred} is used, this means that for each one of the family of models mentioned previously a model will be trained and evaluated using only the premise of the sentence pairs.

\subsection{Stress tests}
To evaluate the robustness of the selected models when training with the dataset, four of the strategies for stress test generation from \cite{naik-etal-2018-stress} are used:
\begin{itemize}
    \item Length mismatch: Make the premise a lot longer than the hypothesis, by adding the expression: \textit{y verdadero es verdadero y verdadero es verdadero y verdadero es verdadero y verdadero es verdadero y verdadero es verdadero}, which does not alter the premise meaning.
    \item Negation: Add negation to the hypothesis without altering the meaning with the expression: \textit{y falso no es verdadero}.
    \item Overlap: Add the expression: \textit{y verdadero es verdadero} to the hypothesis to generate a word mismatch between premise and hypothesis.
    \item Spelling: Misspell a word in the premise.
\end{itemize}
For each strategy a parallel test split is generated from the original test split.

\subsection{Human validation}
To ensure that ESNLIR is useful to other researchers, and given the current budget constraints, a small balanced proportion of 2000 randomly selected pairs from the dataset are annotated by a group of 27 students who are asked to assign a single label to each pair. Only those pairs where the original label matches the majority of human labels are retained. 

Once the gold standard pairs have been extracted, prediction is performed using XLMRoBERTa and classification metrics (f1\_score and accuracy) are applied across the dataset and per genre to determine if the dataset would benefit from human annotation.

\section{Results}

\subsection{Test performance}
\begin{figure}[hbt!]
    \centering
    \includegraphics[width=1\linewidth]{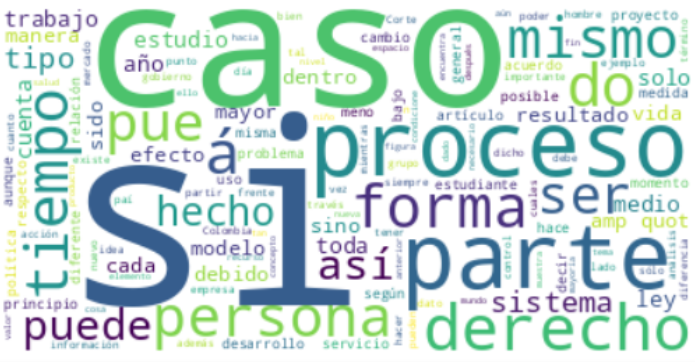}
    \caption{Wordcloud for contrasting class}
    \label{fig:cloud}
\end{figure}

The Table \ref{tab:performance} presents the performance of the models on the test set. When compared to a majority class baseline, the XGBoost model shows a performance increase of 10 points. This suggests that the presence of certain words can indicate the class of a pair. As illustrated in Image \ref{fig:cloud}, words such as \textbf{caso} and \textbf{si} are frequently associated with the \textbf{contrasting} class.

Evaluating BERT-based models reveals that their semantic representations for sentence pairs are superior, nearly doubling the performance of the XGBoost baseline. Notably, XLMRoBERTa, the multilingual model achieves the best overall performance, which could be attributed to the larger size of the dataset used for its pre-training.

\begin{table}[hbt!]
\centering
\small
\begin{tabular}{lll}
\hline
\multicolumn{1}{|l|}{model} & \multicolumn{1}{l|}{accuracy} & \multicolumn{1}{l|}{f1\_score} \\ \hline
Majority class              & 0.25                          & 0.25                           \\
XGBoost                     & 0.35                         & 0.348                          \\
BERTIN                     & 0.663                         & 0.664                          \\
XLMRoBERTa                  & \textbf{0.676}                 & \textbf{0.676}                         
\end{tabular}
\caption{Performance of the baseline models in test}
\label{tab:performance}
\end{table}

In-depth analysis of the classes reveals that for BERT-based models, the 'neutral' class is the simplest to identify. This indicates that distinguishing the lack of semantic relationships between sentences is more straightforward than identifying specific types of semantic relationships. On the other hand, the selection method for this class is different from the others, by not using linking phrases for pair detection, and instead selecting sentences from different paragraphs, which generates a bias that is favourable for the classificaiton of this class. For instance, the following pair is easily identifiable as 'neutral' because there are no shared entities, and the first sentence discusses a \textbf{business association}, whereas the second addresses \textbf{processes for opportunities}:
\begin{itemize}
    \item Premise: 'Dentro de esta definición aparece el concepto de asociaciones de negocios, en las que una organización puede ser miembro de muchas organizaciones virtuales' \textit{(English: Within this definition appears the concept of business partnerships, in which an organization can be a member of many virtual organizations.)}
    \item Hypothesis: 'que no había un proceso formal de localización de las oportunidades, simplemente se cantaban las oportunidades y se “peleaban” las mismas entre los segmentos'  \textit{(English: that there was no formal process for locating opportunities, opportunities were simply sung about and “fought over” between segments)}
\end{itemize}

\begin{table*}[hbt!]
\centering
\begin{tabular}{lllll}
\hline
\multicolumn{1}{|c|}{\textbf{model}} & \multicolumn{1}{c|}{\textbf{contrasting}} & \multicolumn{1}{c|}{\textbf{entailment}} & \multicolumn{1}{c|}{\textbf{neutral}} & \multicolumn{1}{c|}{\textbf{reasoning}} \\ \hline
XGBoost                              & 0.345                                     & 0.436                                    & 0.278                                 & 0.341                                   \\
BERTIN                              & 0.683                                     & 0.626                                    & 0.669                                 & 0.675                                   \\
XLMRoBERTa                           & 0.664                                     & 0.676                                    & 0.695                                 & 0.667                                  
\end{tabular}
\caption{Accuracy per class for baseline models in test}
\end{table*}

\begin{table*}[hbt!]
\centering
\small
\begin{tabular}{lllll}
\hline
\multicolumn{1}{|r|}{\textbf{model}} & \multicolumn{1}{r|}{\textbf{contrasting}} & \multicolumn{1}{r|}{\textbf{entailment}} & \multicolumn{1}{r|}{\textbf{neutral}} & \multicolumn{1}{r|}{\textbf{reasoning}} \\ \hline
XGBoost (both sentences)                              & 0.345                                     & 0.434                                    & 0.287                                 & 0.335                                   \\
BERTIN                              & 0.419                                     & 0.411                                    & 0.261                                 & 0.452                                   \\
XLMRoBERTa                           & 0.441                                     & 0.409                                    & 0.222                                 & 0.485                                  
\end{tabular}
\centering
\caption{Accuracy training and evaluating with only the premise}
\label{tab:performance-premise}
\end{table*}

\begin{table*}[hbt!]
\centering
\small
\begin{tabular}{lllll}
\hline
\multicolumn{1}{|c|}{\textbf{stress test}} & \multicolumn{1}{c|}{\textbf{contrasting}} & \multicolumn{1}{c|}{\textbf{entailment}} & \multicolumn{1}{c|}{\textbf{neutral}} & \multicolumn{1}{c|}{\textbf{reasoning}} \\ \hline
test                                       & 0.664                                     & 0.676                                    & 0.695                                 & 0.667                                   \\
test\_length\_mismatch                     & \textbf{0.513}                            & \textbf{0.524}                           & 0.755                                 & 0.654                                   \\
test\_negation                             & \textbf{0.580}                            & \textbf{0.575}                           & 0.744                                 & 0.656                                   \\
test\_overlap                              & 0.624                                     & 0.637                                    & 0.719                                 & 0.665                                   \\
test\_spelling                             & 0.687                                     & 0.646                                    & 0.677                                 & 0.659                                  
\end{tabular}
\caption{Stress test accuracy per class for XLMRoBERTa}
\label{tab:stress}
\end{table*}

Semantic relationships can be classified into various categories, among which \textbf{contrasting} is the second easiest to identify. This may be the case because contrasting sentences often involve similar entities but contain predicates that are opposites. For instance, consider a pair of sentences where both discuss ongoing research. The first sentence may state that the research reaches the state-of-the-art level, while the second might indicate that additional research is necessary:
\begin{itemize}
    \item Premise: 'cabe resaltar que estos resultados se encuentran de acuerdo con lo que se reporta en la literatura' \textit{(English: It should be noted that these results are in agreement with what is reported in the literature.)}
    \item Hypothesis: 'una investigación más detallada es requerida' \textit{(English: a more detailed investigation is required)} 
\end{itemize}

\begin{figure}
    \centering
    \includegraphics[width=1\linewidth]{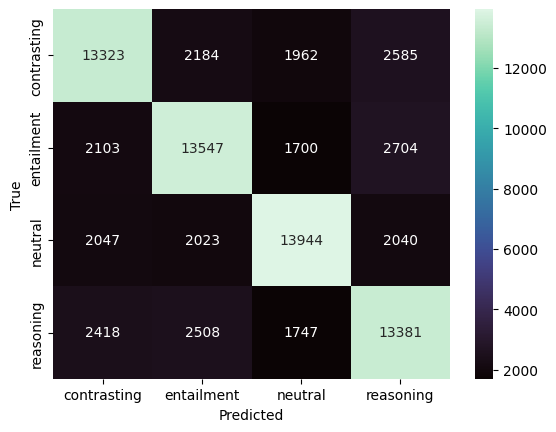}
    \caption{Confusion matrix for XLMRoBERTa}
    \label{fig:conf_mat_xlmr}
\end{figure}

Lastly, the most challenging categories are reasoning and entailment, which models often confuse with each other more than with other categories, as demonstrated in Figure \ref{fig:conf_mat_xlmr}. These confusions may occur because some entailment pairs exhibit specifications or generalizations that might be interpreted as causal relationships, and vice versa. For instance, consider the following two sentences discussing the policies of the Colombian political party UP. The first sentence addresses how these policies failed to consider real-life changes in the country's situation, while the second sentence indicates the party's failure to acknowledge that their policies were not sustainable in the long term:
\begin{itemize}
    \item Premise: 'en particular, el programa de la UP y sus políticas no prestaban atención al tipo de cambio real como determinante de la posición competitiva internacional del país' \textit{(English: In particular, the UP program and its policies did not pay attention to the real exchange rate as a determinant of the country's international competitive position.)}
    \item Hypothesis: 'la UP no reconoció que sus políticas no serían sostenibles en el mediano plazo y que las limitaciones de capacidad se convertirían en un obstáculo insuperable para un rápido crecimiento' \textit{(English: the UP failed to recognize that its policies would not be sustainable in the medium term and that capacity constraints would become an insurmountable obstacle to rapid growth.)}
\end{itemize}
The relationship can be unclear without proper context. For some individuals, the pair relates to \textbf{reasoning} because the failure of the UP to consider real-life changes makes it difficult for them to perceive their policies as unsustainable. For others, the pair is associated with \textbf{entailment} because not acknowledging the unsustainability of the policies can be seen as a result of disregarding the actual changes in the country's situation.

Looking at the details of the correctly classified reasoning pairs, it seems that both sentences have nouns that are related through a process of transformation, which may imply a causal relationship. For example, in the following pair, the first sentence talks about the cremation of a group of people, while the second sentence talks about people collecting their ashes because they believed they were saints. In this case, 'quemados' (burned) is transformed into 'cenizas' (ashes):

\begin{itemize}
    \item Premise: 'esa misma tarde fueron \textbf{quemados}' \textit{(English: that same afternoon they were \textbf{burned})}
    \item Hypothesis: 'la gente quiso coger sus \textbf{cenizas} creyendo que eran las reliquias de unos santos, y por eso, para acabar con el mito, Felipe hizo que tiraran las \textbf{cenizas} al Sena' \textit{(English: people wanted to take their \textbf{ashes} believing that they were the relics of saints, and so, to put an end to the myth, Philip had the \textbf{ashes} thrown into the Seine.)}
\end{itemize}

\subsection{Dataset artifact detection}
In a repeated training and evaluation setup, it was observed that the dataset largely lacks annotation artifacts. As demonstrated in Table \ref{tab:performance-premise}, the accuracy of models trained and evaluated using only the premise is nearly comparable to the XGBoost baseline, yet approximately 25 points lower than the original models. This indicates that both sentences are necessary for the models to effectively identify the type of relationship in a significant portion of the test set. Some performance in this setup may be influenced by unique words within each class, including examples such as misspellings, foreign language words, and entity names:
\begin{itemize}
    \item contrasting: 'tatuajes', 'finiquito', 'imnediata', 'generation', 'exmarido', 'Songkhram'
    \item entailment: 'impresionada', 'Leguízamo', 'Computacional', 'gallinero', 'liviano'
    \item neutral: 'problematizarse', 'palurdos', 'independizados', 'Barbusse', 'Riopaila'
    \item reasoning: 'aeronáuticos', 'primitivamente', 'mitosis', 'larson', 'Marseille', 'Marimbera'
\end{itemize}

\subsection{Stress tests}
The transformations applied to each stress test in Table \ref{tab:stress} affect the classes in different ways. While spelling has minimal impact on performance, the \textbf{contrasting} and \textbf{entailment} classes are significantly influenced by the \textbf{length mismatch} and \textbf{negation} tests. This indicates that sentence length and the presence of negations in the hypothesis substantially affect the discrimination criteria for these classes.

\subsection{Genre performance}

The evaluation of performance by genre, as depicted in Table \ref{tab:genre_performance}, indicates that accuracy does not correspond directly to the number of examples in the datasets. Interestingly, the second-best performing genre lacked any examples in the training split, suggesting that the dataset enables models to generalize to out-of-domain contexts. Nonetheless, the datasets with the poorest performance are associated with non-formal genres, such as \textbf{comments} and \textbf{talks}. This could be because non-formal writing lacks the rigid semantic rules that other genres possess, which limits their potential for generalization.

\begin{table}[hbt!]
\small
\begin{tabular}{lllll}
\hline
\multicolumn{1}{|c|}{\textbf{genre}} & \multicolumn{1}{c|}{\textbf{train|total\_examples}} & \multicolumn{1}{c|}{\textbf{accuracy}} & \multicolumn{1}{c|}{\textbf{f1\_score}} \\ \hline
clinical                             & 0                                                   & 0.735                                  & 0.730                                   \\
legal                                & 253592                                              & 0.727                                  & 0.727                                   \\
books                                & 3021904                                             & 0.684                                  & 0.684                                   \\
articles                             & 254756                                              & 0.679                                  & 0.678                                   \\
theses                               & 269536                                              & 0.674                                  & 0.674                                   \\
news                                 & 877096                                              & 0.671                                  & 0.672                                   \\
comments                             & 100956                                              & 0.647                                  & 0.646                                   \\
talks                                & 0                                                   & 0.597                                  & 0.595                                                              
\end{tabular}
\caption{Performance per genre for XLMRoBERTa}
\label{tab:genre_performance}
\end{table}

\subsection{Out of domain performance}
In Table \ref{tab:out-domain-performance}, it is evident that the highest out-of-sample performance is observed in formal writing genres. In contrast, the lowest performance is noted in non-formal writing genres. This supports the notion that the highly variable semantic rules in these genres negatively impact their performance.

\begin{table}[hbt!]
\resizebox{\columnwidth}{!}{
\begin{tabular}{lll}
\hline
\multicolumn{1}{|c|}{\textbf{corpus}}    & \multicolumn{1}{c|}{\textbf{accuracy}} & \multicolumn{1}{c|}{\textbf{f1\_score}} \\ \hline
esmedical\_\_TEI\_ES                     & 0.744                                  & 0.740                                   \\
esbooks\_\_libreriaunal                  & 0.677                                  & 0.677                                   \\
esbooks\_\_eswikibooks                   & 0.652                                  & 0.653                                   \\
escomments\_\_suicide                    & 0.639                                  & 0.626                                   \\
esmedical\_\_BVS                         & 0.625                                  & 0.598                                   \\
estalks\_\_TED                           & 0.597                                  & 0.595                                   \\
esnews\_\_eswikinews                     & 0.583                                  & 0.592                                   \\
escomments\_\_reddit                     & 0.579                                  & 0.577                                   \\
eslegal\_\_entrevistas\_comision\_verdad & 0.515                                  & 0.513                                  
\end{tabular}
}
\caption{Out-of-domain performance for XLMRoBERTa}
\label{tab:out-domain-performance}
\end{table}

\subsection{Human validation results}
27 students were asked to annotate the 2000 validated examples. To facilitate the annotation process, a LabelStudio instance containing the dataset was made available in the cloud, where each annotator was instructed to select a label for a set. This annotation process resulted in each example having more than one human label, and then only those examples where the original label matched the majority of human labels were retained. This resulted in an unbalanced dataset of 974 examples, with the number of labels shown in tables \ref{tab:class-count-final} and \ref{tab:validate-genre-count}, where all 8 original genres are included. The reduction to less than half the original sample size may mean that for a large part of the dataset, the label defined by the linking phrase does not match the label that a human would use. There are three possible reasons for this, which require further analysis: lack of domain knowledge, lack of clear instructions, or incorrect use of linking phrases at the time the pair source was written.

\begin{table}[!hbt]
\centering
\begin{tabular}{ll}
\hline
\multicolumn{1}{|c|}{\textbf{class}} & \multicolumn{1}{c|}{\textbf{number of examples}} \\ \hline
contrasting                          & 184                                              \\
entailment                           & 219                                              \\
neutral                              & 362                                              \\
reasoning                            & 207                                             
\end{tabular}
\caption{Validated dataset class count}
\label{tab:class-count-final}
\end{table}

\begin{table}[!hbt]
\centering
\begin{tabular}{ll}
\hline
\multicolumn{1}{|l|}{\textbf{genre}} & \multicolumn{1}{c|}{\textbf{number of examples}} \\ \hline
books                                & 217                                              \\
legal                                & 71                                               \\
clinical                             & 51                                               \\
theses                               & 427                                              \\
articles                             & 34                                               \\
comments                             & 67                                               \\
news                                 & 71                                               \\
talks                                & 34                                              
\end{tabular}
\caption{Genre distribution in the validated dataset}
\label{tab:validate-genre-count}
\end{table}

\begin{table*}[hbt!]
\centering
\begin{tabular}{lllll}
\hline
\multicolumn{1}{|c|}{\textbf{genre}} & \multicolumn{1}{l|}{\textbf{original accuracy}} & \multicolumn{1}{c|}{\textbf{validated accuracy}} & \multicolumn{1}{l|}{\textbf{original f1\_score}} & \multicolumn{1}{c|}{\textbf{validated f1\_score}} \\ \hline
\textbf{clinical}                    & 0.735                                           & \textbf{0.780}                                   & 0.730                                            & \textbf{0.776}                                    \\
\textbf{comments}                    & 0.647                                           & \textbf{0.719}                                   & 0.646                                            & 0.708                                             \\
\textbf{books}                       & 0.684                                           & \textbf{0.716}                                   & 0.684                                            & \textbf{0.700}                                    \\
\textbf{articles}                    & 0.679                                           & \textbf{0.697}                                   & \textbf{0.678}                                   & 0.675                                             \\
\textbf{theses}                      & 0.674                                           & \textbf{0.691}                                   & 0.674                                            & \textbf{0.678}                                    \\
\textbf{news}                        & 0.671                                           & \textbf{0.687}                                   & 0.672                                            & \textbf{0.689}                                    \\
\textbf{legal}                       & \textbf{0.727}                                  & 0.667                                            & \textbf{0.727}                                   & 0.610                                             \\
\textbf{talks}                       & 0.597                                           & \textbf{0.647}                                   & 0.595                                            & \textbf{0.630}   
\\\hline 
\textbf{whole dataset}                       & 0.676                                           & \textbf{0.690}                                   & 0.676                                            & \textbf{0.680}
\end{tabular}
\caption{Comparison of metrics between the original test dataset and the human validated test dataset with XLMRoBERTa}
\label{tab:validated-metrics}
\end{table*}

Focusing on the metrics after pruning the dataset, it can be seen in Table \ref{tab:validated-metrics} that the human annotation of the dataset is beneficial because it removes incorrectly labeled pairs and also pairs that are difficult to identify for both models and humans. An example of this is the genre annotations, which went from the second worst to the second best in terms of accuracy, and a possible reason for this is the lack of formal writing and terms that make it easier for a human to classify their examples. On the other hand, when it comes to a more formal domain, such as law, the exact opposite happens, with a decrease in accuracy of 6 points, which means that expertise in the domain is required to make an appropriate labelling. From these kinds of metric trends, it can be concluded that in order to validate this dataset as a whole, it is necessary to select people who have knowledge in the respective domain.

Without taking into account the high percentage of discarded examples, two observations can be made. The first is that the performance after validation is not extremely different from the original test set, meaning that even without full human validation, the dataset can still be useful for training and evaluating relatively good Spanish NLI models. The second is that the overall performance on the dataset, with or without validation, hovers around 68, and based on the same argument found in SciNLI \cite{SCINLI}, the difficulty of classifying the dataset implies that ESNLIR is a challenging benchmark to evaluate the progress of NLU models.

\section{Conclusion}
This paper introduces \textit{ESNLIR}, a new extensive dataset for Spanish multi-genre NLI that incorporates causal relationships between sentences. This dataset demonstrates performance metrics with an accuracy and \textit{f1\_score} exceeding 0.67. While the dataset is robust against artifact annotations, it is sensitive to variations in stress. It also captures specific phenomena associated with each type of semantic relationship; however, it performs less effectively in instances of non-formal writing, such as web comments and talks. Based on these observations, the following future work is proposed:
\begin{itemize}
    \item Evaluating the dataset with more robust models like Large Language Models (LLM).
    \item Intensive annotation of test examples with human annotators to improve the quality of the dataset.
\end{itemize}

\section{Limitations}
Due to our budget constraints, the only human annotation we could pay was the made over a randomly selected small percentage of the test set. However, the results are promising, showing that an annotation over the whole dataset may derive into a larger definite version that is useful for future researchers.

On the other hand, we experimented with more state-of-the-art Large Language Models (LLM), like GPT-4o. But the results were underwhelming when comparing with BERT-based models, and we decided to save these results for another paper in which we carefully explore why they under-perform. Additional to these experiments, we studied the dataset generalization capabilities against other datasets, specifically the Spanish section of XNLI. But similarly to the results with LLM, we expect to publish these results in their own article related to generalization capabilities of NLI models in Spanish. Only XNLI \cite{XNLI} is selected due to its large training size (122k pairs), while INFERES \cite{INFERES} has only 6444 as shown in its huggingface website https://huggingface.co/datasets/venelin/inferes, and the other Spanish NLI datasets are not available on internet.

\bibliography{coling_latex}

\appendix

\section{Appendix}
\label{sec:appendix}
All of the parameters, metrics and models used for BERT models are found in the Zenodo Repository: https://zenodo.org/records/14219405.
\end{document}